%% file: humancm.tex
\documentclass[conference]{IEEEtran}
\IEEEoverridecommandlockouts
\usepackage{cite}
\usepackage{amsmath,amssymb,amsfonts}
\usepackage{algorithm}
\usepackage{algpseudocode}
\usepackage{graphicx}
\usepackage{textcomp}
\usepackage{booktabs}
\usepackage{multirow}
\usepackage{xcolor}
\def\BibTeX{{\rm B\kern-.05em{\sc i\kern-.025em b}\kern-.08em
    T\kern-.1667em\lower.7ex\hbox{E}\kern-.125emX}}
\begin{document}

\title{HumanCM: One Step Human Motion Prediction\\
}

\author{
\IEEEauthorblockN{1\textsuperscript{st} Haojie Liu}
\IEEEauthorblockA{
\textit{School of Engineering Science} \\
\textit{University of Chinese Academy of Sciences}\\
Beijing, China \\
liuhoajie23@mails.ucas.ac.cn}
\and
\IEEEauthorblockN{2\textsuperscript{nd} Suixiang Gao\textsuperscript{*}}
\IEEEauthorblockA{
\textit{School of Mathematical Sciences} \\
\textit{University of Chinese Academy of Sciences}\\
Beijing, China \\
sxgao@ucas.ac.cn\textsuperscript{*}}
\thanks{\textsuperscript{*}Corresponding author}
}

\maketitle

\begin{abstract}
We present \textbf{HumanCM}, a one-step human motion prediction framework built upon consistency models. Instead of relying on multi-step denoising as in diffusion-based methods, HumanCM performs efficient single-step generation by learning a self-consistent mapping between noisy and clean motion states in a latent space. By operating in this compact representation, HumanCM captures long-range temporal dependencies and preserves motion coherence. Experiments on Human3.6M and HumanEva-I demonstrate that HumanCM achieves comparable or superior accuracy to state-of-the-art diffusion models while reducing inference steps by up to two orders of magnitude.

\end{abstract}


\section{Introduction}

Human motion prediction (HMP) is a fundamental task in computer vision and robotics, aiming to forecast future 3D human poses from observed motion sequences. It serves as a core component in applications such as human–robot interaction\cite{10.1007/978-3-319-50115-4_26,LIU2017287,LIU2019272,9281312}, autonomous navigation\cite{7490340, 9093332, 10.1145/3343031.3351082, 9268986}, and immersive virtual environments\cite{yeasin2004multiobject,majoe2009enhanced}. The key challenge lies in accurately modeling the spatial correlations among human joints and the temporal dynamics across frames, while ensuring both high fidelity and real-time responsiveness.

Recent advances in deep generative modeling have greatly enhanced the realism and diversity of motion forecasting. In particular, diffusion-based approaches~\cite{barqueroBeLFusionLatentDiffusion2023, chenHumanMACMaskedMotion2023, weiHumanJointKinematics2023} have shown remarkable success in generating natural and continuous motion trajectories by progressively denoising noisy latent representations. However, their iterative inference—often involving tens or even hundreds of denoising steps—poses a severe computational bottleneck, limiting their applicability in latency-sensitive scenarios such as interactive agents and AR/VR systems.

To overcome this limitation, we propose \textbf{HumanCM}, a one-step human motion prediction framework built upon the recently introduced \emph{Consistency Model (CM)}~\cite{song2023consistencymodels}. Unlike diffusion models that rely on multi-step refinement, consistency models learn a self-consistent mapping between noisy and clean motion states, enabling high-quality generation within a single forward pass. We extend this paradigm to the spatiotemporal domain by integrating temporal embeddings with a Transformer-based architecture, which effectively captures long-range dependencies across both time and body joints. This design allows HumanCM to preserve motion coherence and structural integrity without iterative sampling. 

\begin{figure}[t]
    \centering
    \includegraphics[width=1\linewidth]{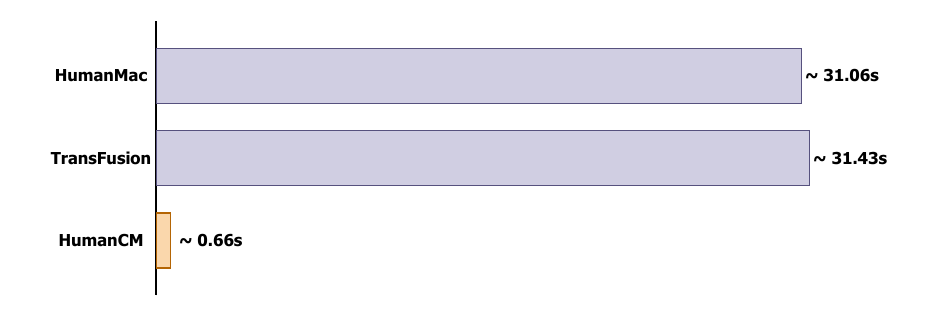}
    \caption{
        Comparison of generation time under identical hardware and batch size.
        HumanCM achieves over two orders of magnitude faster inference compared with diffusion-based models such as HumanMAC and TransFusion.
    }
    \label{fig:time_comparison}
\end{figure}

Furthermore, we enhance consistency learning with a reconstruction-guided objective that stabilizes training and enforces semantic fidelity. Our framework distills diffusion dynamics into a lightweight one-step generator, maintaining competitive accuracy while significantly reducing inference cost. Comprehensive experiments on Human3.6M and HumanEva-I demonstrate that HumanCM achieves state-of-the-art or comparable performance with up to two orders of magnitude faster inference, paving the way for real-time human motion prediction. As illustrated in Figure~\ref{fig:time_comparison}, HumanCM achieves significant acceleration, requiring only a single inference step compared with tens or hundreds of steps in diffusion-based models, resulting in dramatically reduced generation time under identical hardware conditions.

Our main contributions are summarized as follows:
\begin{itemize}
    \item We propose \textbf{HumanCM}, the first consistency-based generative framework for 3D human motion prediction, capable of one-step generation.
    \item We introduce a reconstruction-guided objective that enhances stability and motion fidelity during consistency training.
    \item Extensive experiments demonstrate that HumanCM achieves competitive accuracy with dramatically improved efficiency, enabling real-time motion forecasting.
\end{itemize}

\section{Related Work}

\subsection{Human Motion Prediction}
Early works on Human Motion Prediction (HMP) can be broadly divided into deterministic and stochastic paradigms.  
Deterministic approaches~\cite{acLSTM, DBLP:journals/corr/FragkiadakiLM15, DBLP:journals/corr/MartinezBR17, Corona_2020_CVPR} typically adopt sequence-to-sequence frameworks such as RNNs and CNNs to model temporal evolution, but often suffer from error accumulation and motion discontinuities over long horizons.  
To better capture spatial dependencies among human joints, Graph Convolutional Networks (GCNs)~\cite{mao2019learning, DBLP:journals/corr/abs-2110-04573, Zhong2022SpatioTemporalGG} have been introduced and achieved strong performance.

To model the inherent uncertainty of human motion, stochastic frameworks based on VAEs~\cite{yuanDLowDiversifyingLatent2020, dangDiverseHumanMotion2022} and GANs~\cite{HP-GAN, DeLiGAN} were developed to generate diverse motion trajectories.  
More recently, diffusion-based generative models~\cite{barqueroBeLFusionLatentDiffusion2023, chenHumanMACMaskedMotion2023, weiHumanJointKinematics2023} have achieved state-of-the-art performance by formulating motion prediction as a denoising diffusion process.  
For instance, MotionDiff~\cite{weiHumanJointKinematics2023} interprets motion dynamics as particle diffusion, while HumanMAC~\cite{chenHumanMACMaskedMotion2023} introduces a masked completion mechanism for end-to-end prediction.  
Despite their success, diffusion methods remain computationally expensive due to iterative sampling, motivating the development of faster, more efficient generative alternatives.

\subsection{Diffusion and Consistency Models}
Diffusion probabilistic models~\cite{DDPM, song2021scorebased} have shown exceptional generative quality across image, video, and motion domains but suffer from slow inference.  
Subsequent improvements, such as DDIM~\cite{DDIM}, DPM-Solver~\cite{lu2022dpm}, and progressive distillation~\cite{salimans2022progressive}, reduce sampling steps yet still require multiple iterations.

The recently proposed Consistency Model (CM)~\cite{song2023consistencymodels} offers a new paradigm for fast generative modeling by enforcing prediction consistency across different noise levels.  
Variants including Consistency Distillation (CD), Latent Consistency Models (LCM)~\cite{luo2023latentconsistencymodels}, and SCM~\cite{lu2025simplifyingstabilizingscalingcontinuoustime} extend this concept to latent or continuous-time domains, achieving single-step generation in visual tasks.  
However, their application to spatiotemporal data such as 3D motion remains largely unexplored.  
Our work bridges this gap by adapting the CM framework to the motion domain, introducing spatiotemporal consistency learning that effectively captures both structural and dynamic dependencies while enabling real-time motion generation.

\section{Methodology}
\subsection{Problem Definition}
We represent an observed human motion sequence consisting of $H$ consecutive frames as 
$\mathbf{x}^{(1:H)} = \big[\mathbf{x}^{(1)}, \mathbf{x}^{(2)}, \ldots, \mathbf{x}^{(H)}\big] \in \mathbb{R}^{H \times 3J}$, 
where each $\mathbf{x}^{(h)} \in \mathbb{R}^{3J}$ denotes the 3D joint coordinates of $J$ skeletal joints at frame $h$.  
Given such an observation sequence $\mathbf{x}^{(1:H)}$, the objective of the stochastic human motion prediction (SHMP) task is to infer a set of plausible future motions composed of $F$ frames, denoted as 
$\mathbf{x}^{(H+1:H+F)} = \big[\mathbf{x}^{(H+1)}, \mathbf{x}^{(H+2)}, \ldots, \mathbf{x}^{(H+F)}\big] \in \mathbb{R}^{F \times 3J}$.  
Unlike deterministic forecasting, which yields a single motion trajectory, SHMP aims to model the inherent uncertainty and multimodality of human motion, producing diverse yet physically coherent future sequences conditioned on the observed motion context.

\subsection{Discrete Cosine Transform}

To effectively capture the temporal evolution of human motion, we adopt the Discrete Cosine Transform (DCT) as a compact frequency-domain representation of sequential data.  
Unlike the raw time-domain signals, the DCT encodes motion trajectories through a set of frequency components, allowing the model to represent both rapid transitions and long-term periodic structures in a unified manner.  
This frequency-based representation has been shown to produce smoother, more consistent motion predictions and to facilitate efficient learning of temporal dependencies~\cite{mao2019learning, chenHumanMACMaskedMotion2023, tian2024transfusion}.  

Formally, let $\mathbf{x} \in \mathbb{R}^{(H+F)\times 3J}$ denote a motion sequence composed of $(H+F)$ frames, where each frame contains the 3D joint coordinates of $J$ skeletal joints.  
The sequence is projected into the frequency domain by performing a DCT transformation:
\begin{equation}
\mathbf{y} = \mathrm{DCT}(\mathbf{x}) = D\mathbf{x},
\end{equation}
where $D \in \mathbb{R}^{(H+F)\times(H+F)}$ denotes the predefined orthogonal DCT basis, and $\mathbf{y} \in \mathbb{R}^{(H+F)\times 3J}$ represents the corresponding spectral coefficients.  

Since the DCT is an orthogonal linear transform, the original temporal sequence can be reconstructed exactly by applying its inverse operation:
\begin{equation}
\mathbf{x} = \mathrm{iDCT}(\mathbf{y}) = D^{\top}\mathbf{y}.
\end{equation}
In practice, human motion sequences are generally smooth, meaning that most of their spectral energy resides in the low-frequency bands.  
Hence, instead of preserving all $(H+F)$ frequency components, we retain only the first $L$ rows of $D$ and $D^{\top}$, denoted by $D_{L}$ and $D_{L}^{\top}$, respectively.  
This leads to a compact approximation:
\begin{equation}
\mathbf{y}_{L} = D_{L}\mathbf{x}, \quad \mathbf{x} \approx D_{L}^{\top}\mathbf{y}_{L},
\end{equation}
which suppresses high-frequency noise while preserving the essential temporal dynamics and global motion structure.

\subsection{HumanCM}

\begin{figure*}[t]
    \centering
    \includegraphics[width=\linewidth]{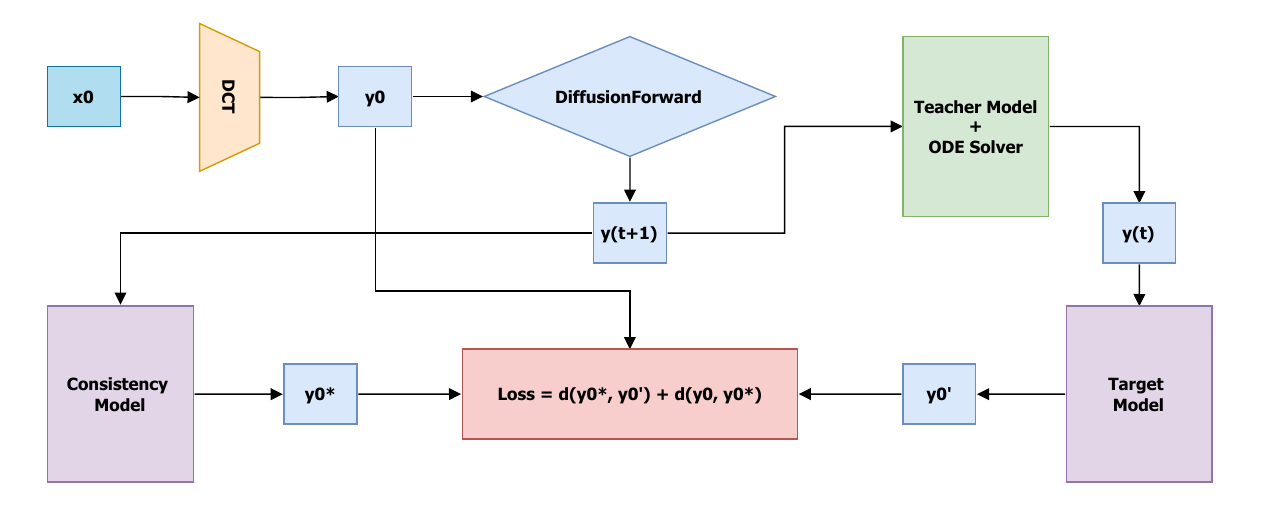}
    \caption{Overall training pipeline of \textbf{HumanCM}. The model learns a self-consistent mapping between noisy and clean motion representations under the consistency distillation framework.}
    \label{fig:train_pipeline}
\end{figure*}

The Consistency Model (CM)~\cite{song2023consistencymodels} is an efficient generative framework enabling one-step or few-step sampling.  
Given a Probability Flow ODE (PF-ODE) that gradually perturbs data into noise, CM learns a mapping $f:(\mathbf{x}_t,t)\mapsto\mathbf{x}_\epsilon$ satisfying the \emph{self-consistency condition}:
\begin{equation}
f(\mathbf{x}_t,t)=f(\mathbf{x}_{t'},t'), \quad \forall\, t,t'\in[\epsilon,T],
\label{eq:self_consistency}
\end{equation}
ensuring identical outputs along the PF-ODE trajectory.  
The parameterized model $f_{\Theta}$ is defined as
\begin{equation}
f_{\Theta}(\mathbf{x},t)=c_{\text{skip}}(t)\mathbf{x}+c_{\text{out}}(t)\mathbf{F}_{\Theta}(\mathbf{x},t),
\label{eq:param}
\end{equation}
where $\mathbf{F}_{\Theta}$ denotes a neural network and $c_{\text{skip}}(\epsilon)=1$, $c_{\text{out}}(\epsilon)=0$.  
CMs are typically trained through \emph{Consistency Distillation (CD)} with loss:
\begin{equation}
\mathcal{L}(\Theta,\Theta^{-};\Phi)=\mathbb{E}\big[d\big(f_{\Theta}(\mathbf{x}_{t_{n+1}},t_{n+1}),\,f_{\Theta^{-}}(\hat{\mathbf{x}}^{\Phi}_{t_n},t_n)\big)\big],
\label{eq:cm_loss}
\end{equation}
where $f_{\Theta^{-}}$ is the EMA target, and $\hat{\mathbf{x}}^{\Phi}_{t_n}$ is a PF-ODE update:
\begin{equation}
\hat{\mathbf{x}}^{\Phi}_{t_n} = \mathbf{x}_{t_{n+1}}+(t_n-t_{n+1})\,\Phi(\mathbf{x}_{t_{n+1}},t_{n+1},\varnothing).
\label{eq:phi}
\end{equation}

\begin{algorithm*}[ht]
\caption{HumanCM Training}
\label{alg:train}
\begin{algorithmic}[1]
\Require dataset $\mathcal{D}_y=\{(\mathbf{y}_0,c)\}$, encoder $\mathcal{E}$, solver $\Phi$, EMA rate $\rho$
\State Initialize model parameters $\Theta$, target parameters $\Theta^{-}\!\leftarrow\!\Theta$
\For{each iteration}
    \State Sample $(\mathbf{y}_0,c)\!\sim\!\mathcal{D}_y$, skip interval $k$, and guidance scale $w\!\sim\!\mathcal{U}[w_{\min},w_{\max}]$
    \State Sample $t_{n+k}>t_n$ and perturb $\mathbf{y}_{t_{n+k}}\!\leftarrow\!\text{DiffusionForward}(\mathbf{y}_0,t_{n+k})$
    \State Compute $\hat{\mathbf{y}}_{t_n}^{\Phi,w} \leftarrow \mathbf{y}_{t_{n+k}} +(1+w)\Phi(\mathbf{y}_{t_{n+k}},t_{n+k},t_n,c)
-w\Phi(\mathbf{y}_{t_{n+k}},t_{n+k},t_n,\varnothing)$
    \State Evaluate:
    \[
    \mathcal{L} = d(f_{\Theta}(\mathbf{y}_{t_{n+1}},t_{n+1},w,c),f_{\Theta^{-}}(\hat{\mathbf{y}}^{\Phi}_{t_n},t_n,w,c))
    + \lambda\, d(f_{\Theta}(\mathbf{y}_t,t,w^*,c),\mathbf{y}_0)
    \]
    \State Update $\Theta \leftarrow \text{Adam}(\nabla_{\Theta}\mathcal{L})$
    \State Update EMA target: $\Theta^{-}\leftarrow\rho\,\Theta^{-}+(1-\rho)\,\Theta$
\EndFor
\end{algorithmic}
\end{algorithm*}

The Latent Consistency Model (LCM)~\cite{luo2023latentconsistencymodels} extends CM into the latent domain $\mathcal{D}_y=\{(\mathbf{y},c)\,|\,\mathbf{y}=\mathcal{E}(\mathbf{x})\}$, where $\mathcal{E}$ is a pretrained encoder and $c$ is an optional condition.  
LCMs adopt a discrete-time schedule~\cite{DDPM,lu2022dpm} and enforce consistency across $k$-step intervals $(t_{n+k}\!\to\!t_n)$, significantly accelerating convergence.  
To incorporate classifier-free guidance (CFG)~\cite{ho2022classifier}, the update rule becomes:
\begin{equation}
\begin{aligned}
\hat{\mathbf{y}}_{t_n} = &\mathbf{y}_{t_{n+k}} +(1+w)\Phi(\mathbf{y}_{t_{n+k}},t_{n+k},t_n,c) \\
&-w\Phi(\mathbf{y}_{t_{n+k}},t_{n+k},t_n,\varnothing),
\end{aligned}
\label{eq:lcm_cfg}
\end{equation}
where $w\!\sim\!\mathcal{U}[w_{\min},w_{\max}]$.  
The final mapping is $f:(\mathbf{y}_t,t,w,c)\mapsto\mathbf{y}_0$, following MotionLCM~\cite{motionlcm}.  

To achieve more stable and physically coherent motion generation, we augment the standard consistency loss with a reconstruction constraint that directly anchors predictions to the ground truth data manifold.  
Specifically, the total objective combines consistency and reconstruction terms:
\begin{equation}
\begin{aligned}
\mathcal{L} = &\mathbb{E}\big[d(f_{\Theta}(\mathbf{y}_{t_{n+1}},t_{n+1},w,c),f_{\Theta^{-}}(\hat{\mathbf{y}}^{\Phi}_{t_n},t_n,w,c))\big] \\
&+ \lambda\,\mathbb{E}\big[d(f_{\Theta}(\mathbf{y}_t,t,w^*,c),\mathbf{y}_0)\big],
\end{aligned}
\end{equation}
where $w^*$ is a fixed guidance scale and $\lambda$ balances the two objectives.  
The first term enforces temporal self-consistency, while the second encourages the model to maintain semantic fidelity and motion realism in the latent space.  

\begin{figure}[t]
    \centering
    \includegraphics[width=\linewidth]{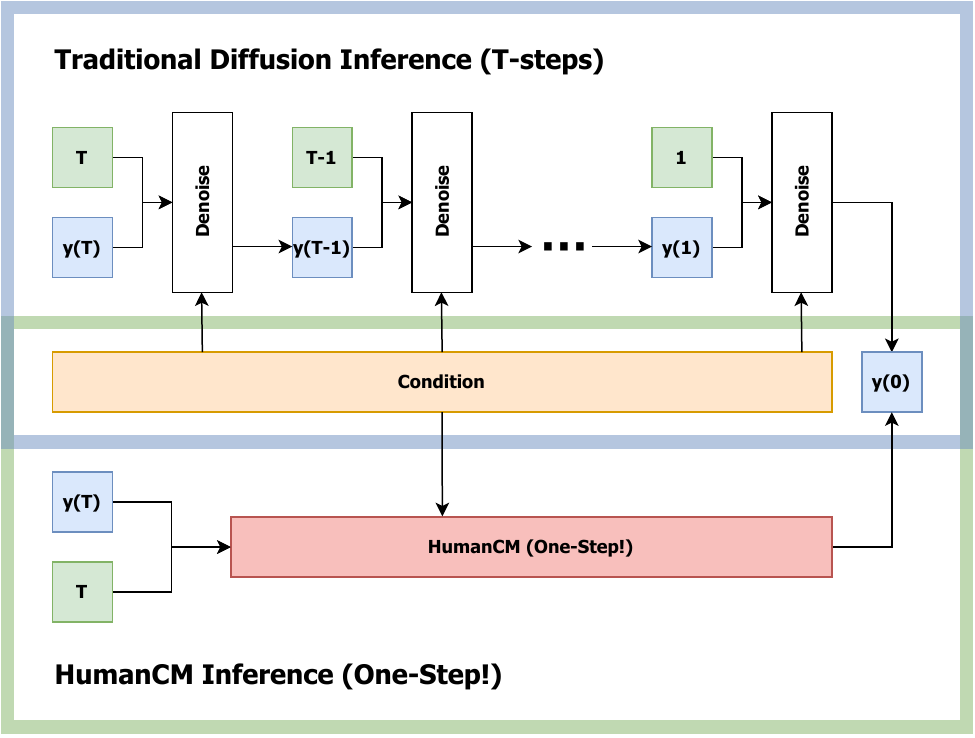}
    \caption{
    \textbf{Timeline comparison of motion generation processes.}
    The blue box illustrates the \textit{traditional diffusion-based inference}, 
    where the model progressively denoises from $y(T)$ to $y(0)$ through $T$ iterative steps.
    In contrast, the green box highlights our \textbf{HumanCM one-step inference}, 
    which directly maps the noisy input $y(T)$ to the clean motion state $y(0)$ in a single forward pass, 
    achieving over two orders of magnitude faster generation.
    }
    \label{fig:denoising_timeline}
\end{figure}

The overall training procedure is illustrated in Figure \ref{fig:train_pipeline}, while the detailed training and inference steps are provided in Algorithm \ref{alg:train} and Algorithm \ref{alg:infer}, respectively. As shown in Figure~\ref{fig:denoising_timeline}, HumanCM eliminates the multi-step denoising required by diffusion-based models (blue box) and performs direct one-step generation (green box), offering significant acceleration without compromising motion quality. Note that the self-attention operations are performed over latent temporal tokens after frequency transformation, rather than directly on raw joint coordinates; thus, visualizing attention maps on the spatial domain is not physically meaningful.

\begin{algorithm}[t]
\caption{HumanCM Inference (One-step Generation)}
\label{alg:infer}
\begin{algorithmic}[1]
\Require condition $c$, initial noise $\mathbf{y}_T$, trained model $f_{\Theta}$, guidance scale $w^*$
\State $\mathbf{y}_{0} \leftarrow f_{\Theta}(\mathbf{y}_T,T,w^*,c)$
\State Decode motion: $\hat{\mathbf{x}}_0 = \mathcal{D}(\mathbf{y}_0)$
\State \Return generated motion sequence $\hat{\mathbf{x}}_0$
\end{algorithmic}
\end{algorithm}

\section{Experiments}

\input{tables/table1}
\input{tables/table2}

\subsection{Datasets and Metrics}

We conduct comprehensive evaluations on two widely used benchmarks for human motion prediction, namely Human3.6M~\cite{Human36M} and HumanEva-I~\cite{humaneva}. The assessment is performed using five standard quantitative metrics.

Prediction accuracy is evaluated through Average Displacement Error (ADE) and Final Displacement Error (FDE). Specifically, ADE represents the average positional deviation across the entire sequence, while FDE focuses on the deviation at the final frame.  
For the multi-modal prediction setting, we further employ Multi-Modal ADE (MMADE) and Multi-Modal FDE (MMFDE), which extend ADE and FDE by clustering future trajectories conditioned on similar historical contexts. 

\subsection{Implementation details}

We train \emph{HumanCM} by applying consistency distillation in a pre-trained DDIM model from Transfusion\cite{tian2024transfusion} with the same model structure. Experiments are conducted with PyTorch and one NVIDIA A100-SXM4-40GB GPU. Training is performed with the Adam optimizer, an initial learning rate of 0.0003 decayed by 0.9 every 75 epochs, for 5000 epochs both on Human3.6M and HumanEva-I. The training guidance scale range $[w_{\min},w_{\max}] = [0,1]$ and the infer guidance scale is  $w^* = 1/125$. The scale $\lambda$ for reconstruction constraint is set as $1/15$. To capture long-range temporal dependencies, HumanCM employs temporal embeddings with a fixed dimension of $d_t$, which is chosen to match the model’s latent dimension (e.g., 256), ensuring compatibility with the Transformer’s multi-head attention layers.

\subsection{Results}

Table~\ref{table:quant_results} presents the quantitative comparisons on Human3.6M and HumanEva-I datasets. HumanCM consistently achieves superior performance across most evaluation metrics. On Human3.6M, our model attains the lowest ADE (0.382) and FDE (0.504), outperforming previous state-of-the-art methods such as GSPS~\cite{maoGeneratingSmoothPose2021} and MotionDiff~\cite{weiHumanJointKinematics2023}. Although the MMFDE shows a slight gap compared to GSPS, HumanCM demonstrates a strong balance between short-term accuracy (ADE) and long-term trajectory coherence (FDE), indicating its stable motion prediction capability.

On HumanEva-I, HumanCM again achieves the best ADE (0.231), surpassing the diffusion-based MotionDiff (0.232) while requiring significantly fewer sampling steps. The improvement in ADE and competitive results in FDE (0.304) verify that our approach preserves temporal consistency and spatial plausibility in motion prediction. Overall, HumanCM yields state-of-the-art or comparable results while maintaining higher computational efficiency.

As shown in Table~\ref{tab:steps_comparison}, existing diffusion-based models, including MotionDiff~\cite{weiHumanJointKinematics2023}, HumanMAC~\cite{chenHumanMACMaskedMotion2023}, and TransFusion~\cite{tian2024transfusion}, require 10–100 iterative sampling steps to generate motion sequences. In contrast, HumanCM leverages a consistency-based framework capable of one-step generation. Despite this drastic reduction in sampling steps, HumanCM still achieves competitive or superior quantitative performance. This demonstrates that consistency training effectively distills the diffusion process into a single-step mapping, resulting in both faster inference and more stable trajectory generation.

\section{Conclusion}

In this paper, we presented \textbf{HumanCM}, a novel consistency-based framework for one-step human motion prediction. By distilling diffusion dynamics into a single-step mapping, HumanCM eliminates the computational burden of iterative denoising while maintaining high-quality and temporally coherent motion generation. The model integrates temporal embeddings with a self-attention based spatial-temporal encoder, which effectively captures both long-range temporal dependencies and inter-joint interactions critical for realistic motion forecasting.

Comprehensive experiments on Human3.6M and HumanEva-I datasets show that HumanCM achieves comparable or superior accuracy to diffusion-based approaches, while reducing inference steps by up to two orders of magnitude. These results highlight the potential of consistency models as a powerful alternative to traditional diffusion frameworks for spatiotemporal generation tasks.

Future work will explore extending the HumanCM framework toward controllable and interactive motion synthesis, multi-agent coordination, and cross-modal applications integrating vision and text. We believe this research opens new directions for efficient, real-time human motion generation in next-generation AI-driven systems.

\bibliographystyle{unsrt}  
\bibliography{references}

\end{document}

%% file: tables/table1.tex
\begin{table*}[ht]
\caption{Quantitative results on Human3.6M and HumanEva-I. The best results are highlighted in \textbf{bold}.}
\centering
\resizebox{\textwidth}{!}{%
\begin{tabular}{lc|cccc|cccc}
\toprule
\multirow{2}{*}{Method} & \multirow{2}{*}{\# Loss} & \multicolumn{4}{c|}{Human3.6M} & \multicolumn{4}{c}{HumanEva-I} \\ 
 &  & ADE$\downarrow$ & FDE$\downarrow$ & MMADE$\downarrow$ & MMFDE$\downarrow$ & ADE$\downarrow$ & FDE$\downarrow$ & MMADE$\downarrow$ & MMFDE$\downarrow$ \\ 
\midrule
DeLiGAN\cite{DeLiGAN} & 1 & 0.483 & 0.534 & 0.520 & 0.545 & 0.306 & 0.322 & 0.385 & 0.371 \\
MT-VAE\cite{MT-VAE} & 3 & 0.457 & 0.595 & 0.716 & 0.883 & 0.345 & 0.403 & 0.518 & 0.577 \\
BoM\cite{BoM} & 1 & 0.448 & 0.533 & 0.514 & 0.544 & 0.271 & 0.279 & 0.373 & 0.351 \\
DSF\cite{DSF} & 2 & 0.493 & 0.592 & 0.550 & 0.599 & 0.273 & 0.290 & 0.364 & 0.340 \\
DLow\cite{yuanDLowDiversifyingLatent2020} & 3 & 0.425 & 0.518 & 0.495 & 0.531 & 0.233 & 0.243 & 0.343 & 0.331  \\
GSPS\cite{maoGeneratingSmoothPose2021} & 5 & 0.389 & 0.496 & \textbf{0.476} & \textbf{0.525} & 0.233 & 0.244 & \textbf{0.343} & 0.331 \\
MOJO\cite{MOJO} & 3 & 0.412 & 0.514 & 0.497 & 0.538 & 0.234 & 0.244 & 0.369 & 0.347 \\
MotionDiff\cite{weiHumanJointKinematics2023} & 4 & 0.411 & 0.509 & 0.508 & 0.536 & 0.232 & \textbf{0.236} & 0.352 & \textbf{0.320} \\ 

\midrule
HumanCM & 2 & \textbf{0.382} & \textbf{0.504} & 0.517 & 0.570 & \textbf{0.231} & 0.304 & 0.448 & 0.507 \\ 
\bottomrule
\end{tabular}
}
\label{table:quant_results}
\end{table*}

%% file: tables/table2.tex
\begin{table}[htb]
\caption{Comparison of sampling steps among different motion generation models.}
\centering
\begin{tabular}{lcc}
\toprule
\textbf{Model} & \textbf{Type} & \textbf{Sampling Steps} \\
\midrule
BeLFusion~\cite{barqueroBeLFusionLatentDiffusion2023} & Diffusion-based & 10 \\
MotionDiff~\cite{weiHumanJointKinematics2023} & Diffusion-based & 100 \\
HumanMAC~\cite{chenHumanMACMaskedMotion2023} & Diffusion-based & 100 \\
TransFusion~\cite{tian2024transfusion} & Diffusion-based & 100 \\
\midrule
\textbf{HumanCM (Ours)} & Consistency-based & \textbf{1} \\
\bottomrule
\end{tabular}
\label{tab:steps_comparison}
\end{table}

%% file: humancm.bbl
\begin{thebibliography}{10}

\bibitem{10.1007/978-3-319-50115-4_26}
Ozgur~S. Oguz, Volker Gabler, Gerold Huber, Zhehua Zhou, and Dirk Wollherr.
\newblock Hybrid human motion prediction for action selection within human-robot collaboration.
\newblock In Dana Kuli{\'{c}}, Yoshihiko Nakamura, Oussama Khatib, and Gentiane Venture, editors, {\em 2016 International Symposium on Experimental Robotics}, pages 289--298, Cham, 2017. Springer International Publishing.

\bibitem{LIU2017287}
Hongyi Liu and Lihui Wang.
\newblock Human motion prediction for human-robot collaboration.
\newblock {\em Journal of Manufacturing Systems}, 44:287--294, 2017.
\newblock Special Issue on Latest advancements in manufacturing systems at NAMRC 45.

\bibitem{LIU2019272}
Zitong Liu, Quan Liu, Wenjun Xu, Zhihao Liu, Zude Zhou, and Jie Chen.
\newblock Deep learning-based human motion prediction considering context awareness for human-robot collaboration in manufacturing.
\newblock {\em Procedia CIRP}, 83:272--278, 2019.
\newblock 11th CIRP Conference on Industrial Product-Service Systems.

\bibitem{9281312}
Ruixuan Liu and Changliu Liu.
\newblock Human motion prediction using adaptable recurrent neural networks and inverse kinematics.
\newblock {\em IEEE Control Systems Letters}, 5(5):1651--1656, 2021.

\bibitem{7490340}
Brian Paden, Michal Čáp, Sze~Zheng Yong, Dmitry Yershov, and Emilio Frazzoli.
\newblock A survey of motion planning and control techniques for self-driving urban vehicles.
\newblock {\em IEEE Transactions on Intelligent Vehicles}, 1(1):33--55, 2016.

\bibitem{9093332}
Nemanja Djuric, Vladan Radosavljevic, Henggang Cui, Thi Nguyen, Fang-Chieh Chou, Tsung-Han Lin, Nitin Singh, and Jeff Schneider.
\newblock Uncertainty-aware short-term motion prediction of traffic actors for autonomous driving.
\newblock In {\em 2020 IEEE Winter Conference on Applications of Computer Vision (WACV)}, pages 2084--2093, 2020.

\bibitem{10.1145/3343031.3351082}
Shiming Ge, Shengwei Zhao, Xindi Gao, and Jia Li.
\newblock Fewer-shots and lower-resolutions: Towards ultrafast face recognition in the wild.
\newblock In {\em Proceedings of the 27th ACM International Conference on Multimedia}, MM '19, page 229–237, New York, NY, USA, 2019. Association for Computing Machinery.

\bibitem{9268986}
Zan Gao, Leming Guo, Weili Guan, An-An Liu, Tongwei Ren, and Shengyong Chen.
\newblock A pairwise attentive adversarial spatiotemporal network for cross-domain few-shot action recognition-r2.
\newblock {\em IEEE Transactions on Image Processing}, 30:767--782, 2021.

\bibitem{yeasin2004multiobject}
Mohammed Yeasin, Ediz Polat, and Rajeev Sharma.
\newblock A multiobject tracking framework for interactive multimedia applications.
\newblock {\em IEEE transactions on multimedia}, 6(3):398--405, 2004.

\bibitem{majoe2009enhanced}
Dennis Majoe, Lars Widmer, and Juerg Gutknecht.
\newblock Enhanced motion interaction for multimedia applications.
\newblock In {\em Proceedings of the 7th International Conference on Advances in Mobile Computing and Multimedia}, pages 13--19, 2009.

\bibitem{barqueroBeLFusionLatentDiffusion2023}
German Barquero, Sergio Escalera, and Cristina Palmero.
\newblock {{BeLFusion}}: {{Latent Diffusion}} for {{Behavior-Driven Human Motion Prediction}}.
\newblock In {\em 2023 {{IEEE}}/{{CVF International Conference}} on {{Computer Vision}} ({{ICCV}})}, pages 2317--2327, Paris, France, October 2023. IEEE.

\bibitem{chenHumanMACMaskedMotion2023}
Ling-Hao Chen, Jiawei Zhang, Yewen Li, Yiren Pang, Xiaobo Xia, and Tongliang Liu.
\newblock {{HumanMAC}}: {{Masked Motion Completion}} for {{Human Motion Prediction}}.
\newblock In {\em Proceedings of the {{IEEE}}/{{CVF International Conference}} on {{Computer Vision}}}. arXiv, August 2023.

\bibitem{weiHumanJointKinematics2023}
Dong Wei, Huaijiang Sun, Bin Li, Jianfeng Lu, Weiqing Li, Xiaoning Sun, and Shengxiang Hu.
\newblock Human {{Joint Kinematics Diffusion-Refinement}} for {{Stochastic Motion Prediction}}.
\newblock {\em Proceedings of the AAAI Conference on Artificial Intelligence}, 37(5):6110--6118, June 2023.

\bibitem{song2023consistencymodels}
Yang Song, Prafulla Dhariwal, Mark Chen, and Ilya Sutskever.
\newblock Consistency models, 2023.

\bibitem{acLSTM}
Zimo Li, Yi~Zhou, Shuangjiu Xiao, Chong He, and Hao Li.
\newblock Auto-conditioned {LSTM} network for extended complex human motion synthesis.
\newblock {\em CoRR}, abs/1707.05363, 2017.

\bibitem{DBLP:journals/corr/FragkiadakiLM15}
Katerina Fragkiadaki, Sergey Levine, and Jitendra Malik.
\newblock Recurrent network models for kinematic tracking.
\newblock {\em CoRR}, abs/1508.00271, 2015.

\bibitem{DBLP:journals/corr/MartinezBR17}
Julieta Martinez, Michael~J. Black, and Javier Romero.
\newblock On human motion prediction using recurrent neural networks.
\newblock {\em CoRR}, abs/1705.02445, 2017.

\bibitem{Corona_2020_CVPR}
Enric Corona, Albert Pumarola, Guillem Alenya, and Francesc Moreno-Noguer.
\newblock Context-aware human motion prediction.
\newblock In {\em Proceedings of the IEEE/CVF Conference on Computer Vision and Pattern Recognition (CVPR)}, June 2020.

\bibitem{mao2019learning}
Wei Mao, Miaomiao Liu, Mathieu Salzmann, and Hongdong Li.
\newblock Learning trajectory dependencies for human motion prediction.
\newblock In {\em Proceedings of the IEEE/CVF international conference on computer vision}, pages 9489--9497, 2019.

\bibitem{DBLP:journals/corr/abs-2110-04573}
Theodoros Sofianos, Alessio Sampieri, Luca Franco, and Fabio Galasso.
\newblock Space-time-separable graph convolutional network for pose forecasting.
\newblock {\em CoRR}, abs/2110.04573, 2021.

\bibitem{Zhong2022SpatioTemporalGG}
Chongyang Zhong, Lei Hu, Zihao Zhang, Yongjing Ye, and Shi hong Xia.
\newblock Spatio-temporal gating-adjacency gcn for human motion prediction.
\newblock {\em 2022 IEEE/CVF Conference on Computer Vision and Pattern Recognition (CVPR)}, pages 6437--6446, 2022.

\bibitem{yuanDLowDiversifyingLatent2020}
Ye~Yuan and Kris Kitani.
\newblock {{DLow}}: {{Diversifying Latent Flows}} for {{Diverse Human Motion Prediction}}.
\newblock In {\em {{ECCV}}}. arXiv, July 2020.

\bibitem{dangDiverseHumanMotion2022}
Lingwei Dang, Yongwei Nie, Chengjiang Long, Qing Zhang, and Guiqing Li.
\newblock Diverse {{Human Motion Prediction}} via {{Gumbel-Softmax Sampling}} from an {{Auxiliary Space}}.
\newblock In {\em Proceedings of the 30th {{ACM International Conference}} on {{Multimedia}}}. arXiv, July 2022.

\bibitem{HP-GAN}
Emad Barsoum, John~R. Kender, and Zicheng Liu.
\newblock {HP-GAN:} probabilistic 3d human motion prediction via {GAN}.
\newblock {\em CoRR}, abs/1711.09561, 2017.

\bibitem{DeLiGAN}
Swaminathan Gurumurthy, Ravi~Kiran Sarvadevabhatla, and Venkatesh~Babu Radhakrishnan.
\newblock Deligan : Generative adversarial networks for diverse and limited data.
\newblock {\em CoRR}, abs/1706.02071, 2017.

\bibitem{DDPM}
Jonathan Ho, Ajay Jain, and Pieter Abbeel.
\newblock Denoising diffusion probabilistic models.
\newblock {\em CoRR}, abs/2006.11239, 2020.

\bibitem{song2021scorebased}
Yang Song, Jascha Sohl-Dickstein, Diederik~P Kingma, Abhishek Kumar, Stefano Ermon, and Ben Poole.
\newblock Score-based generative modeling through stochastic differential equations.
\newblock In {\em International Conference on Learning Representations}, 2021.

\bibitem{DDIM}
Jiaming Song, Chenlin Meng, and Stefano Ermon.
\newblock Denoising diffusion implicit models.
\newblock {\em arXiv:2010.02502}, October 2020.

\bibitem{lu2022dpm}
Cheng Lu, Yuhao Zhou, Fan Bao, Jianfei Chen, Chongxuan Li, and Jun Zhu.
\newblock Dpm-solver: A fast ode solver for diffusion probabilistic model sampling in around 10 steps.
\newblock {\em Advances in neural information processing systems}, 35:5775--5787, 2022.

\bibitem{salimans2022progressive}
Tim Salimans and Jonathan Ho.
\newblock Progressive distillation for fast sampling of diffusion models.
\newblock {\em arXiv preprint arXiv:2202.00512}, 2022.

\bibitem{luo2023latentconsistencymodels}
Simian Luo, Yiqin Tan, Longbo Huang, Jian Li, and Hang Zhao.
\newblock Latent consistency models: Synthesizing high-resolution images with few-step inference, 2023.

\bibitem{lu2025simplifyingstabilizingscalingcontinuoustime}
Cheng Lu and Yang Song.
\newblock Simplifying, stabilizing and scaling continuous-time consistency models, 2025.

\bibitem{tian2024transfusion}
Sibo Tian, Minghui Zheng, and Xiao Liang.
\newblock Transfusion: A practical and effective transformer-based diffusion model for 3d human motion prediction.
\newblock {\em IEEE Robotics and Automation Letters}, 9(7):6232--6239, 2024.

\bibitem{ho2022classifier}
Jonathan Ho and Tim Salimans.
\newblock Classifier-free diffusion guidance.
\newblock {\em arXiv preprint arXiv:2207.12598}, 2022.

\bibitem{motionlcm}
Wenxun Dai, Ling-Hao Chen, Jingbo Wang, Jinpeng Liu, Bo~Dai, and Yansong Tang.
\newblock Motionlcm: Real-time controllable motion generation via latent consistency model.
\newblock In {\em ECCV}, pages 390--408, 2025.

\bibitem{MT-VAE}
Xinchen Yan, Akash Rastogi, Ruben Villegas, Kalyan Sunkavalli, Eli Shechtman, Sunil Hadap, Ersin Yumer, and Honglak Lee.
\newblock {{MT-VAE}}: {{Learning}} motion transformations to generate multimodal human dynamics.
\newblock In Vittorio Ferrari, Martial Hebert, Cristian Sminchisescu, and Yair Weiss, editors, {\em Computer Vision -- {{ECCV}} 2018}, pages 276--293, Cham, 2018. Springer International Publishing.

\bibitem{BoM}
Apratim Bhattacharyya, Bernt Schiele, and Mario Fritz.
\newblock Accurate and diverse sampling of sequences based on a "best of many" sample objective.
\newblock {\em CoRR}, abs/1806.07772, 2018.

\bibitem{DSF}
Ye~Yuan and Kris Kitani.
\newblock Diverse trajectory forecasting with determinantal point processes.
\newblock {\em CoRR}, abs/1907.04967, 2019.

\bibitem{maoGeneratingSmoothPose2021}
Wei Mao, Miaomiao Liu, and Mathieu Salzmann.
\newblock Generating {{Smooth Pose Sequences}} for {{Diverse Human Motion Prediction}}.
\newblock In {\em 2021 {{IEEE}}/{{CVF International Conference}} on {{Computer Vision}} ({{ICCV}})}, pages 13289--13298, Montreal, QC, Canada, October 2021. IEEE.

\bibitem{MOJO}
Yan Zhang, Michael~J. Black, and Siyu Tang.
\newblock We are {{More}} than {{Our Joints}}: {{Predicting}} how {{3D Bodies Move}}.
\newblock In {\em 2021 {{IEEE}}/{{CVF Conference}} on {{Computer Vision}} and {{Pattern Recognition}} ({{CVPR}})}, pages 3371--3381, Nashville, TN, USA, June 2021. IEEE.

\bibitem{Human36M}
Catalin Ionescu, Dragos Papava, Vlad Olaru, and Cristian Sminchisescu.
\newblock Human3.6m: Large scale datasets and predictive methods for 3d human sensing in natural environments.
\newblock {\em IEEE Transactions on Pattern Analysis and Machine Intelligence}, 36(7):1325--1339, 2014.

\bibitem{humaneva}
Leonid Sigal, Alexandru Balan, and Michael Black.
\newblock Humaneva: Synchronized video and motion capture dataset and baseline algorithm for evaluation of articulated human motion.
\newblock {\em International Journal of Computer Vision}, 87:4--27, 03 2010.

\end{thebibliography}
